\documentclass[sigconf]{acmart}
\usepackage{booktabs}
\usepackage{graphicx}
\usepackage{subfigure}
\usepackage{multirow}
\usepackage{mathtools}
\usepackage{bbding}
\usepackage{amsmath}

\usepackage{amssymb}
\usepackage{mathtools}
\usepackage{amsthm}
\usepackage{adjustbox}
\usepackage{booktabs}
\usepackage[table]{xcolor}
\usepackage{multirow}
\usepackage{graphicx}
\usepackage{mathtools}
\usepackage{bbding}
\usepackage{makecell}
\usepackage{subfigure}

\definecolor{base}{RGB}{255,0,0}
\definecolor{rl1}{RGB}{94,162,247}
\definecolor{rl2}{RGB}{104,187,47}
\definecolor{rl3}{RGB}{213,164,25}
\definecolor{sft1}{RGB}{178,40,178}
\definecolor{sft2}{RGB}{85,78,140}

\AtBeginDocument{%
  }

\setcopyright{acmlicensed}
\copyrightyear{2026}
\acmYear{2026}
\setcopyright{cc}
\setcctype{by}
\acmConference[KDD '26]{Proceedings of the 32nd ACM SIGKDD Conference on Knowledge Discovery and Data Mining V.2}{August 09--13, 2026}{Jeju Island, Republic of Korea}
\acmBooktitle{Proceedings of the 32nd ACM SIGKDD Conference on Knowledge Discovery and Data Mining V.2 (KDD '26), August 09--13, 2026, Jeju Island, Republic of Korea}
\acmDOI{10.1145/3770855.3817679}
\acmISBN{979-8-4007-2259-2/2026/08}
\begin{document}

\title{Sparsity Curse: Understanding RLVR Model Parameter Space from Model Merging}

\author{Chenrui Wu}
\authornote{Equal contribution.}
\orcid{0000-0002-8349-2682}
\affiliation{%
\institution{Zhejiang University}
\city{Hangzhou}
\state{}
\country{China}}
\affiliation{%
\institution{Simon Fraser University}
\city{Burnaby}
\state{}
\country{Canada}}
\email{chenrui_wu@sfu.ca}

\author{Zexi Li}
\authornotemark[1]
\orcid{0000-0003-0831-3549}
\affiliation{%
\institution{Knowin AI}
\city{Shenzhen}
\state{}
\country{China}}
\affiliation{
\institution{The Chinese University of Hong Kong}
\city{Hong Kong}
\state{}
\country{China}}
\email{lizexi@knowin.ai}

\author{Jiajun Bu}
\orcid{0000-0002-1097-2044}
\affiliation{%
\department{Zhejiang Key Lab of Accessible Perception and Intelligent Systems}
\institution{Zhejiang University}
\city{Hangzhou}
\state{}
\country{China}}
\email{bjj@zju.edu.cn}

\author{Jiangchuan Liu}
\orcid{0000-0001-6592-1984}
\affiliation{%
\department{School of Computing Science}
\institution{Simon Fraser University}
\city{Burnaby}
\state{}
\country{Canada}}
\email{jcliu@sfu.ca}

\author{Haishuai Wang}
\orcid{0000-0003-1617-0920}
\authornote{Corresponding author: Haishuai Wang.}
\affiliation{%
\department{College of Computer Science}
\institution{Zhejiang University}
\city{Hangzhou}
\state{}
\country{China}}
\email{haishuai.wang@zju.edu.cn}

\renewcommand{\shortauthors}{Chenrui Wu, Zexi Li, Jiajun Bu, Jiangchuan Liu, and Haishuai Wang}

\begin{abstract}
Reinforcement Learning with Verifiable Reward (RLVR) has emerged as a powerful post-training paradigm that surpasses Supervised Fine-Tuning (SFT) in eliciting reasoning intelligence and resisting catastrophic forgetting. Recent studies further reveal that RLVR induces highly sparse and off-principal parameter updates compared to SFT. This naturally raises the question: does such sparsity make RLVR models more amenable to model merging? If so, model merging would offer a scalable, training-free path to aggregate diverse reasoning capabilities from independently trained RLVR models. Surprisingly, we find the opposite, uncovering a \emph{sparsity curse}: the sparse RLVR updates are spread farther apart in parameter space, forming near-orthogonal shortcuts that make aggregation inherently fragile. This is likely rooted in the stochasticity of RL optimization and the diversity of emergent reasoning patterns. Unlike SFT models that converge to shared, flat basins and merge naturally, RLVR models suffer severe degradation under standard merging methods. Through systematic empirical analysis of the update geometry, we characterize the mechanisms behind this failure and propose Sensitivity-aware Resolving Merging (SAR-Merging), a merging recipe tailored for the unique structure of RLVR parameter spaces. SAR-Merging resolves conflicts in overlapping update regions via Fisher Information-based sensitivity arbitration, followed by magnitude-aware sparsification and rescaling to preserve fragile reasoning pathways. Experiments on mathematical and coding benchmarks demonstrate that SAR-Merging substantially outperforms existing merging methods on RLVR models, enabling both single-task enhancement and multi-capability fusion.
\end{abstract}



\maketitle

\section{Introduction}

\begin{figure}[t]
    \centering
\includegraphics[width=1\linewidth]{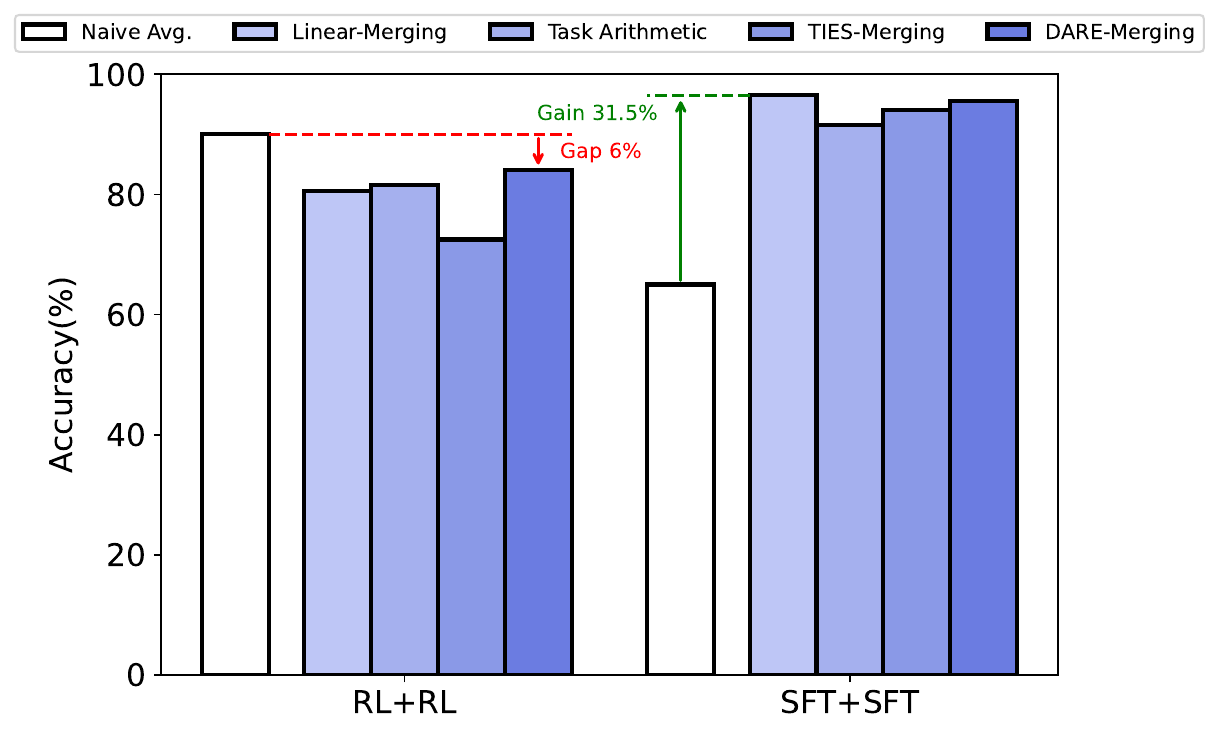}
    \caption{The performance gain and drop of merging two 7B RLVR/SFT models with GSM8K data. The naive avg. means the average of the individual accuracy of the two parent models.}
    \label{fig:simplemerge}
\end{figure}
Reinforcement Learning with Verifiable Reward (RLVR)~\cite{GRPO,jin2025rl,zhang2026onpolicy} has become the dominant post-training paradigm for large language models (LLMs), powering reasoning breakthroughs in models such as OpenAI-o3~\cite{jaech2024openai} and DeepSeek-R1~\cite{GRPO}. Compared to Supervised Fine-Tuning (SFT)~\cite{radford2018improving,li2021prefix,hu2022lora}, RLVR better elicits multi-step reasoning and resists catastrophic forgetting~\cite{grattafiori2024llama,Qwen2.5-Math,jiang2023mistral7b,zhang2026onpolicy}. Recent analysis by Zhu et al.~\cite{zhu2025path} further reveals a striking property of RLVR in parameter space: unlike SFT, which updates along the principal singular directions of the weight matrices, RLVR updates are \emph{sparse} and \emph{off-principal}, modifying only a small fraction of parameters in low-curvature subspaces while preserving the pretrained spectral structure. This finding, also supported by Mukherjee et al.~\cite{mukherjee2025reinforcement}, shows that RLVR and SFT operate in fundamentally different optimization regimes.

This distinction naturally raises a question: does the sparse, off-principal nature of RLVR updates make these models more amenable to \emph{model merging}? Model merging~\cite{modelsoup,TaskArithmetic,ties,DARE,acm,aim,RobustMerge,matena2022merging,li2025improving} combines independently trained models into a single unified model without additional training. If RLVR's sparse updates are compatible across models, merging would offer a scalable, training-free path to aggregate diverse reasoning capabilities from separately trained RLVR models.

Surprisingly, we find the opposite. As shown in Figure~\ref{fig:simplemerge}, standard merging methods that work well for SFT models cause severe performance degradation when applied to RLVR models. We attribute this to a \emph{sparsity curse}: the sparse, off-principal updates learned by independent RLVR models are spread far apart in parameter space, forming near-orthogonal shortcuts that make aggregation inherently fragile. This fragility likely stems from the stochasticity of RL optimization and the diversity of emergent reasoning patterns across independent training runs. While RAM-Merging~\cite{ram} has also observed degradation in RL model merging, their experiments use relatively dense RL agent models whose update patterns resemble SFT, and their approach does not generalize to the highly sparse RLVR models studied here.

In this paper, we systematically analyze the mechanisms behind this failure through empirical studies of update sparsity, layer-wise orthogonality, and activation density geometry. Based on these analyses, we propose Sensitivity-aware Resolving Merging (SAR-Merging), a merging recipe tailored for the unique structure of RLVR parameter spaces. SAR-Merging first resolves conflicts in overlapping update regions using Fisher Information to preserve the more task-sensitive parameters. It then applies magnitude-aware sparsification and rescaling to the merged task vectors, respecting the intrinsic sparsity of RLVR and preventing the dense collapse that destroys reasoning pathways. We summarize our contributions as follows:
\begin{itemize}
    \item We uncover the \emph{sparsity curse} in RLVR model merging and provide systematic empirical analysis of how RLVR's sparse, off-principal updates differ from SFT in the context of model aggregation.
    \item We propose SAR-Merging, a model merging method tailored for RLVR models, combining sensitivity-based conflict resolution with sparsity-preserving rescaling.
    \item Extensive experiments on mathematical and coding benchmarks demonstrate that SAR-Merging substantially outperforms existing methods, enabling both single-task enhancement and multi-capability fusion for RLVR models.
\end{itemize}

\section{Related Works}
\subsection{Post-training of LLMs}
There are two main techniques for LLM post-training: Supervised Fine-Tuning (SFT) and Reinforcement Learning (RL). An LLM with the capability of next-token prediction often fails to follow user instructions precisely. To bridge this gap, SFT has been widely adopted to align the model's output distribution with user intent~\cite{dodge2020fine,zhao2023survey}. SFT uses a small amount of labeled data to adjust model parameters for a specific task~\cite{radford2018improving,li2021prefix,hu2022lora}. While effective, SFT is limited by the quality of demonstration data and often suffers from exposure bias.

To further align models with human values, Reinforcement Learning from Human Feedback (RLHF)~\cite{rlhf} has emerged as a standard paradigm. Proximal Policy Optimization (PPO)~\cite{PPO} maximizes the expected reward while maintaining a KL-divergence constraint to the reference model. Direct Preference Optimization (DPO)~\cite{DPO} derives a closed-form solution for the optimal policy using a simple classification loss. More recently, Reinforcement Learning with Verifiable Reward (RLVR) has gained prominence as a post-training strategy for reasoning tasks~\cite{GRPO,jaech2024openai}, like math and coding. Unlike RLHF, which relies on learned reward models, RLVR uses objective, rule-based verification signals (e.g., checking mathematical correctness) to provide rewards. Group Relative Policy Optimization (GRPO)~\cite{GRPO} uses group advantages as reward signals, making RLVR applicable to many domains ~\cite{intentrl,gui_g1,tian2026seea}. This paradigm has powered state-of-the-art reasoning models such as DeepSeek-R1~\cite{GRPO} and OpenAI-o3~\cite{jaech2024openai}. Notably, Zhu et al.~\cite{zhu2025path} show that RLVR operates in a fundamentally different optimization regime from SFT: its parameter updates are sparse and off-principal, concentrating in low-curvature subspaces rather than along the principal singular directions favored by SFT.

\subsection{Model Merging}
Model merging facilitates training-free integration by merging multiple task-specific models into a unified framework, which is widely adopted in different areas, e.g., federated learning~\cite{fedlaw,fedma}, model editing~\cite{wang2024wise,ke_mm}, and multi-task learning~\cite{TaskArithmetic,yang2024adamerging}. A foundational approach, Linear Averaging~\cite{modelsoup}, computes a simple weighted mean of the model parameters. Task Arithmetic~\cite{TaskArithmetic} proposes task vectors, defined as the deviation from the pre-trained weights, which are subsequently aggregated and added back to the base model. TIES-Merging~\cite{ties} refines task arithmetic by sparsifying these vectors and employing a sign consensus algorithm, thereby preserving the distinct capabilities of individual models. DARE~\cite{DARE} enables the efficient fusion of multi-task fine-tuned models through stochastic pruning and rescaling, also combined with TIES-Merging. Beyond parameter-level operations, AIM~\cite{aim} leverages activation space statistics to identify and preserve critical weights in the pre-trained model. ACM~\cite{acm} also exploits the mutual information between the model's activations and the base model's to guide layer-wise weighted aggregation. There are also some domain-specific model merging paradigms designed for LoRA merging~\cite{RobustMerge,OSRM,knots,LoRA-LEGO,K-Merge}, continual merging~\cite{MINGLE,dop,opcm,NUFILT}, and so on. However, these methods generally target SFT models and exhibit performance degradation when applied to RLVR models. RAM-Merging~\cite{ram} first considers the merging of RLVR agent models. However, the agent models they adopted have relatively low update sparsity, similar to SFT models, which do not generalize to the highly sparse RLVR models.\looseness=-1

\section{Preliminaries}
\subsection{Supervised Fine-Tuning (SFT)}
We consider an autoregressive language model parameterized by $\theta$, which defines a conditional distribution over output sequences $y = (y_1,\ldots,y_T)$ given an input prompt $x$. The model likelihood factorizes as
\begin{equation}
p_\theta(y \mid x) \;=\; \prod_{t=1}^{T} p_\theta(y_t \mid x, y_{<t}).
\end{equation}
In supervised fine-tuning, the model is trained on a dataset of human annotations
$\mathcal{D} = \{(x^{(i)}, y^{(i)})\}_{i=1}^N$.
The objective is to maximize the log-likelihood of the reference outputs, or equivalently to minimize the negative log-likelihood loss:
\begin{equation}
\mathcal{L}_{\mathrm{SFT}}(\theta)
\;=\;
-\mathbb{E}_{(x,y)\sim \mathcal{D}}
\Big[
\sum_{t=1}^{|y|}
\log p_\theta(y_t \mid x, y_{<t})
\Big].
\end{equation}

This training procedure relies on teacher forcing, where the ground-truth prefix $y_{<t}$ is provided at each step. SFT encourages the model to imitate the conditional distribution implicit in the dataset and typically yields stable optimization and strong language modeling performance. However, it is limited by the quality and coverage of labeled outputs and does not directly optimize for abstract or non-local objectives such as human preferences.

\subsection{Reinforcement Learning with Verifiable Reward (RLVR)}
\setlength{\tabcolsep}{20pt}
\begin{table*}[!t]
\centering
\caption{\textbf{Analysis of parameter update sparsity in various RL algorithms using open-source models from Hugging Face.} The models are in the most widely adopted bfloat16 data type, and the threshold of update is $1\times10^{-5}$.}
\label{tab:sparsity_joint}
\resizebox{0.8\textwidth}{!}{%
\begin{tabular}{@{} l l l l r @{}}
\toprule
\textbf{Base Model} & \textbf{Finetuned (FT) Model} & \textbf{Algorithm} & \textbf{Data} & \textbf{$\mathrm{Sparsity}_{\mathrm{bf16}}$} \\
\midrule

\href{https://huggingface.co/Qwen/Qwen2.5-Math-1.5B}{Qwen-1.5B} &
\href{https://huggingface.co/deepseek-ai/DeepSeek-R1-Distill-Qwen-1.5B}{DS-R1-Distill-Qwen-1.5B} &
SFT & Mixed & 2.8\% \\

\href{https://huggingface.co/deepseek-ai/DeepSeek-R1-Distill-Qwen-1.5B}{DS-R1-Distill-Qwen-1.5B} &
\href{https://huggingface.co/ziyan98/lul-sft}{lul-SFT} &
SFT & Math & 12.06\% \\

\href{https://huggingface.co/deepseek-ai/DeepSeek-R1-Distill-Qwen-1.5B}{DS-R1-Distill-Qwen-1.5B} &
\href{https://huggingface.co/eoet9r/MiniMath}{MiniMath} &
SFT & Math & 60.39\%  \\

\href{https://huggingface.co/deepseek-ai/DeepSeek-R1-Distill-Qwen-1.5B}{DS-R1-Distill-Qwen-1.5B} &
\href{https://huggingface.co/agentica-org/DeepCoder-1.5B-Preview}{DeepCoder-1.5B-Preview} &
GRPO & Code & 70.30\% \\

\href{https://huggingface.co/deepseek-ai/DeepSeek-R1-Distill-Qwen-1.5B}{DS-R1-Distill-Qwen-1.5B} &
\href{https://huggingface.co/agentica-org/DeepScaleR-1.5B-Preview}{DeepScaleR-1.5B-Preview} &
GRPO & Math & 77.36\% \\

\href{https://huggingface.co/deepseek-ai/DeepSeek-R1-Distill-Qwen-1.5B}{DS-R1-Distill-Qwen-1.5B} &
\href{https://huggingface.co/RLinf/RLinf-math-1.5B}{RLinf-math-1.5B} &
GRPO & Math & 70.90\% \\

\href{https://huggingface.co/deepseek-ai/DeepSeek-R1-Distill-Qwen-1.5B}{DS-R1-Distill-Qwen-1.5B} &
\href{https://huggingface.co/Salesforce/E1-Math-1.5B}{E1-Math-1.5B} &
GRPO & Math &75.74\% \\

\midrule
\href{https://huggingface.co/Qwen/Qwen2.5-7B}{Qwen2.5-7B} &
\href{https://huggingface.co/Qwen/Qwen2.5-Math-7B}{Qwen2.5-Math-7B} &
SFT & Math & 2.3\% \\

\href{https://huggingface.co/Qwen/Qwen2.5-Math-7B}{Qwen2.5-Math-7B}  &
\href{https://huggingface.co/PhysicsWallahAI/Aryabhata-1.0}{Aryabhata-1.0} &
SFT & Math &  3.2\% \\

\href{https://huggingface.co/Qwen/Qwen2.5-Math-7B}{Qwen2.5-Math-7B}  &
\href{https://huggingface.co/Satori-reasoning/Satori-SFT-7B}{Satori-SFT-7B} &
SFT & Math &  21.84\% \\

\href{https://huggingface.co/Qwen/Qwen2.5-Math-7B}{Qwen2.5-Math-7B}  &
\href{https://huggingface.co/muriloluz/Qwen-2.5-7B-MATH-RL}{Qwen-2.5-7B-MATH-RL} &
GRPO & Math &  85.70\% \\

\href{https://huggingface.co/Qwen/Qwen2.5-Math-7B}{Qwen2.5-Math-7B}  &
\href{https://huggingface.co/sail/Qwen2.5-Math-7B-Oat-Zero}{Qwen2.5-Math-7B-Oat-Zero} &
DRPO & Math &  76.70\% \\

\href{https://huggingface.co/Qwen/Qwen2.5-Math-7B}{Qwen2.5-Math-7B} &
\href{https://huggingface.co/GD-ML/Qwen2.5-Math-7B-GPG}{Qwen2.5-Math-7B-GPG} &
GPG & Math &  46.53\% \\

\bottomrule
\end{tabular}}
\end{table*}

RLVR frames the language model as a stochastic policy $p_\theta(y \mid x)$ that generates a full response $y$ conditioned on a prompt $x$. Instead of matching a reference output, the model is optimized to maximize a scalar verifiable reward $r(x,y)$ that evaluates the correctness or quality of the generated sequence. In RLVR, this reward is provided by an objective verifier, such as an answer checker, a rule-based evaluator, or executable tests, rather than a learned reward model. The expected return objective is:
\begin{equation}
J(\theta)
\;=\;
\mathbb{E}_{x \sim \mathcal{D}_x}
\Big[
\mathbb{E}_{y \sim p_\theta(\cdot \mid x)}
\big[
r(x,y)
\big]
\Big],
\end{equation}
where $\mathcal{D}_x$ denotes the distribution of prompts.

The gradient of this objective can be estimated using policy gradient methods:
\begin{equation}
\nabla_\theta J(\theta)
\;=\;
\mathbb{E}_{x \sim \mathcal{D}_x,\; y \sim p_\theta(\cdot \mid x)}
\big[
r(x,y)\,\nabla_\theta \log p_\theta(y \mid x)
\big].
\end{equation}
A baseline $b(x)$ is often introduced to reduce the variance of the gradient estimator:
\begin{equation}
\nabla_\theta J(\theta)
\;\approx\;
\mathbb{E}_{x \sim \mathcal{D}_x,\; y \sim p_\theta(\cdot \mid x)}
\big[
(r(x,y) - b(x))\,\nabla_\theta \log p_\theta(y \mid x)
\big].
\end{equation}

To prevent the optimized policy from drifting too far from a reference model $p_{\theta_0}$, a KL-divergence penalty is commonly added:
\begin{equation}
\widetilde{J}(\theta)
\;=\;
\mathbb{E}_{x y \sim p_\theta}
\big[
r(x,y)
\big]
\;- \;
\beta\,
\mathbb{E}_{x }
\big[
\mathrm{KL}\big(p_\theta(\cdot \mid x)\,\|\,p_{\theta_0}(\cdot \mid x)\big)
\big],
\end{equation}
where $\beta$ controls the strength of regularization. Optimization is typically carried out using stable policy optimization algorithms such as PPO~\cite{PPO}, DPO~\cite{DPO}, GRPO~\cite{GRPO}, and DAPO~\cite{yu2026dapo}.

\subsection{Task Vectors}
Let $\theta^0 \in \mathbb{R}^d$ denote a base model, and let $\theta^t$ be a model $\theta^0$ fine-tuned on task~$t$ by SFT or RL. The corresponding \emph{task vector} is defined as the parameter difference as:
\begin{equation}
\delta^t\;=\; \theta^t - \theta^0,
\end{equation}
which represents the task-induced update in parameter space. Task vectors provide a linearized abstraction of model adaptation and enable post-hoc model editing without additional training.

Given a scaling coefficient $\alpha$, task-specific behavior can be injected via:
\begin{equation}
\theta \;=\; \theta^0 + \alpha \delta^t.
\end{equation}
More generally, multiple task vectors can be combined to form a merged model:
\begin{equation}
\theta_{\mathrm{merge}}
\;=\;
\theta^0 + \sum_{t} \alpha \delta^t.
\end{equation}

This approach assumes approximate linearity of task-relevant updates and compatibility among task vectors, allowing modular composition of behaviors learned from separate fine-tuning procedures.

\section{Empirical Study of RLVR Model Merging}
\subsection{Observation on RLVR Model Merging}

\begin{figure*}[ht]
    \centering
     \subfigure[RLVR model layer-wise sparsity]{
    \includegraphics[width=0.45\linewidth]{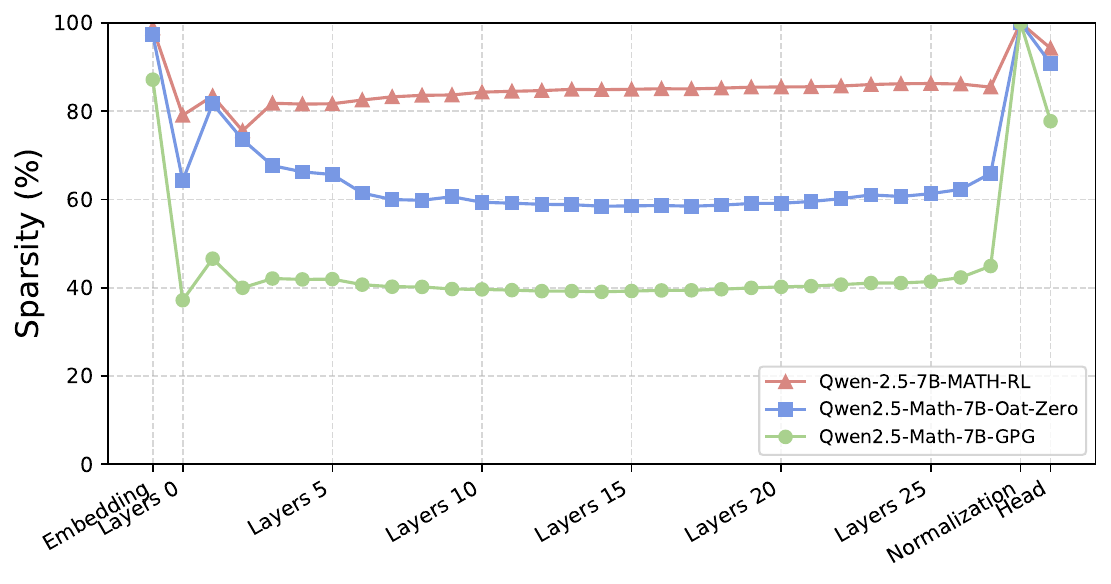}
     \label{fig:RLSparse}}
        \subfigure[SFT model layer-wise sparsity]{
    \includegraphics[width=0.45\linewidth]{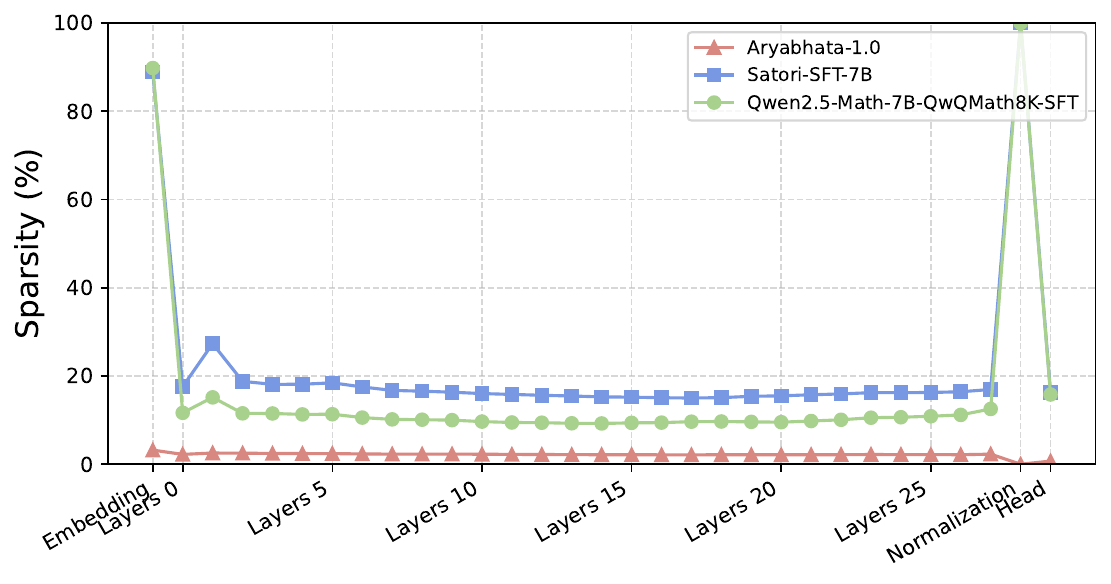}
     \label{fig:SFTSparse}}
    \caption{Layer-wise update sparsity profiles of 7B RLVR and SFT models.}
    \label{fig:layer_sparsity}
\end{figure*}
Most previous methods, including Linear average~\cite{modelsoup}, Task Arithmetic~\cite{TaskArithmetic}, TIES-merging~\cite{ties}, and DARE~\cite{DARE}, were developed under the assumption of SFT parameter updates. When we attempted to merge the RL models with the same approaches, we encountered unexpected performance degradation. As shown in Figure~\ref{fig:simplemerge}, we combined two RL models and two SFT models using classic methods. We test the 7B models in Table~\ref{tab:sparsity_joint} under GSM8K questions. Although the original accuracy of the two SFT models is not high (naive average in Figure~\ref{fig:simplemerge}), any merging strategy could significantly improve the performance of the merged model. Conversely, for the RL models, although the two individual models perform well on average, applying any merging approach resulted in a performance decrease. Therefore, we attempt to observe and dissect RL training from multiple perspectives to understand the RLVR process and find an effective merging method.

\subsection{Sparsity of RLVR}
\textbf{RL updates are sparse.} Given the base model $\theta^0$ and fine-tuned model $\theta^t$, we have the sparsity $(\theta^0,\theta^t) \coloneqq 1-||\theta^t- \theta^0||_0/n$, where $\|\cdot\|_0$ denotes the size of non-zero updates based on the threshold value, and $n$ is the total parameter dimension. The sparsity reflects how many parameters are not updated, where a lower value means a denser update.

Reinforcement learning updates have been found to be sparse in recent works~\cite{mukherjee2025reinforcement,zhu2025path}. Compared to SFT, reinforcement learning updates modify fewer parameters. As shown in Table~\ref{tab:sparsity_joint}, we analyzed two sets of models of different sizes, which were updated using SFT and RLVR based on their respective base models. At the commonly used bfloat16 accuracy, the RLVR model has fewer updated parameters (1-sparsity), regardless of the RL algorithm used. Previous work~\cite{mukherjee2025reinforcement} has further analyzed the updated subnets, which we will not repeat here.

\textbf{Layer-wise sparsity.} Beyond the sparsity of the entire model, we further analyzed the update degree of each layer in Figure~\ref{fig:layer_sparsity}. For RL models, most updates occur in the intermediate layers. For the token embedding layer and head layer, updates are infrequent, while the normalization layer is almost never updated. The updated intermediate layers maintain a relatively stable update magnitude, independent of layer depth. For the SFT model, updates are also concentrated in the intermediate layers. The token embedding layer and normalization layer are almost never updated. Unlike RLVR, the SFT's output head is also significantly updated to facilitate label supervision. Chen et al.~\cite{chen2026does} also show that RL can improve the activation intensity of critical paths within the model. RL focuses updates on a few key neurons to generate high-intensity activation, which aligns with our hierarchical sparse updates in Figure~\ref{fig:layer_sparsity}.
\begin{figure}[!t]
    \centering
     \subfigure[1.5B RLVR models]{
    \includegraphics[width=1\columnwidth]{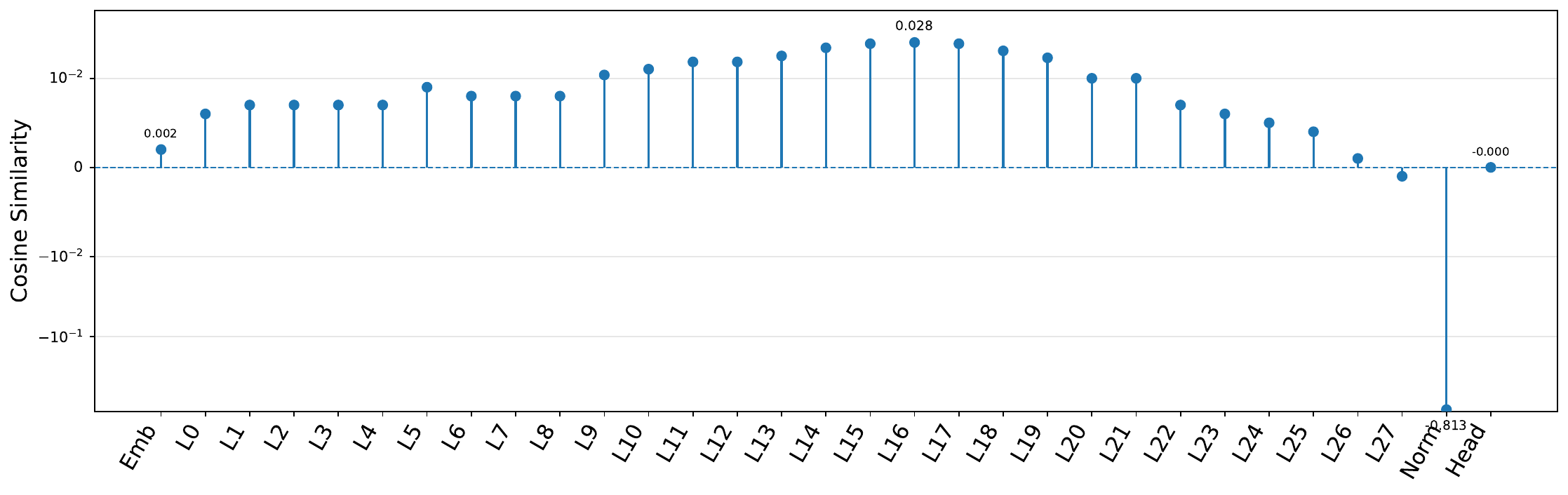}
     \label{fig:1.5BRLsim}}
     \subfigure[7B RLVR models]{
    \includegraphics[width=1\columnwidth]{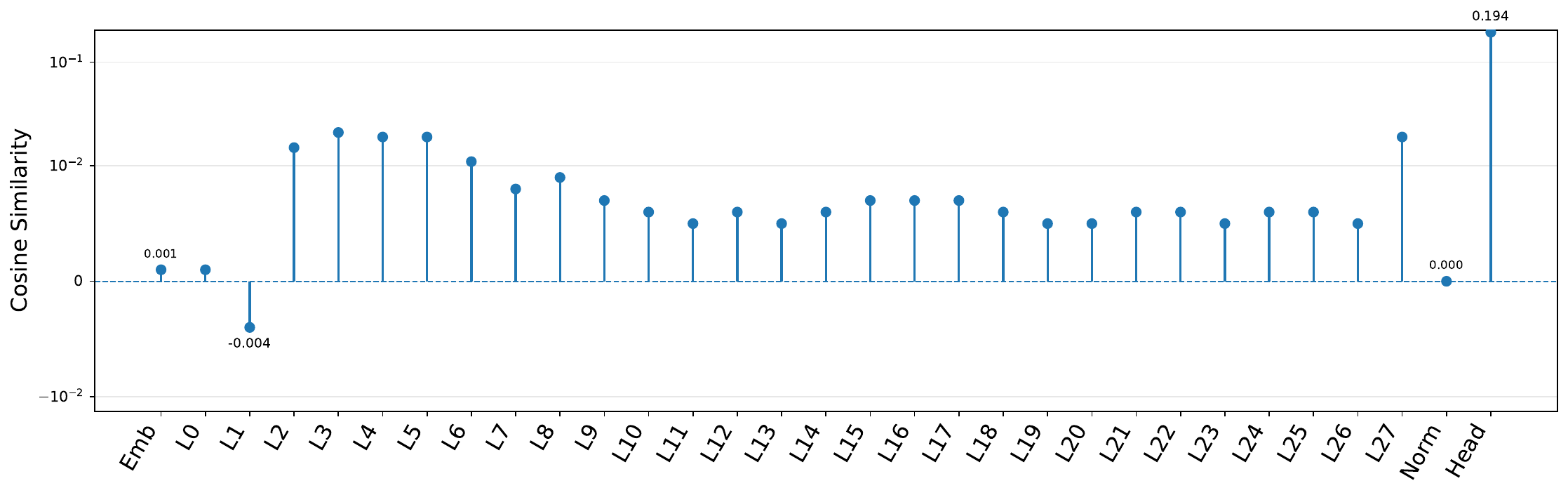}
     \label{fig:7BRLsim}}
       \vspace{-15pt}
    \caption{ Layer-wise cosine similarity analysis of weight updates. 1.5B models: DeepScaleR-1.5B-Preview vs. RLinf-math-1.5B. 7B models: Qwen-2.5-7B-MATH-RL vs. Qwen2.5-Math-7B-Oat-Zero.}
    \label{fig:similarity}
  \vspace{-15pt}
\end{figure}

\begin{figure}[!t]
    \centering
     \subfigure[Update dynamics of 1B model]{
    \includegraphics[width=0.45\columnwidth]{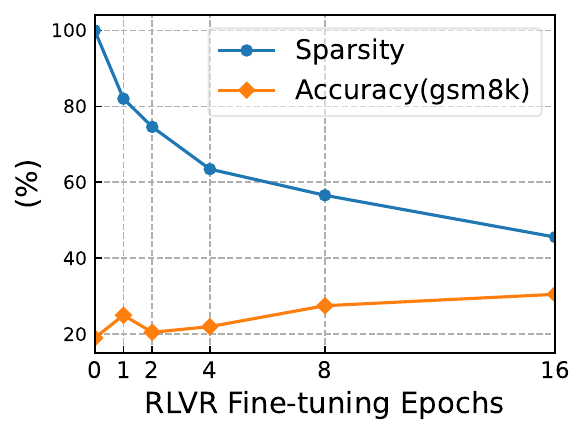}
     \label{fig:dynamic1B}}
         \subfigure[Update dynamics of 4B model]{
    \includegraphics[width=0.45\columnwidth]{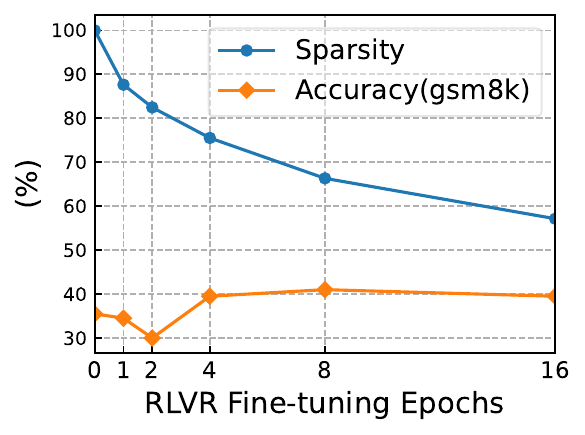}
     \label{fig:dynamic4B}}
\vspace{-10pt}
    \caption{\textbf{Update sparsity and accuracy of RLVR model under PPO algorithm from checkpoints in EvolLM~\cite{qi2025evolm}.} }
    \label{fig:dynamic}
\vspace{-5pt}
\end{figure}

\textbf{Layer-wise update similarity.} We further analyzed the cosine similarity of parameter updates between two RL models at each layer. The results in Figure~\ref{fig:similarity} show that the updates of two RL models at the same layer are nearly all orthogonal. In other words, different RL models randomly seek a shortcut in the parameter space, instead of converging along shared principal directions as SFT models do. This phenomenon is similar to the findings in~\cite{zhang2025reinforcement}, which indicates that RL increases activation diversity by enabling different inputs to activate distinct functional sub-paths. Thus, independent RL models reinforce different orthogonal circuit shortcuts for the same task.

\textbf{RL training dynamics.} Using checkpoints from EvolLM~\cite{qi2025evolm} (1B and 4B models), we track how sparsity evolves during PPO training in Figure~\ref{fig:dynamic}. As training progresses, update sparsity gradually decreases as RL expands its update footprint. However, accuracy does not increase monotonically but fluctuates, consistent with RL exploring diverse shortcuts rather than following a single optimization path.

\subsection{Update Landscape of RLVR}
\begin{figure*}[!t]
    \centering
     \subfigure[RL activation density (2D)]{
    \includegraphics[width=0.5\linewidth]{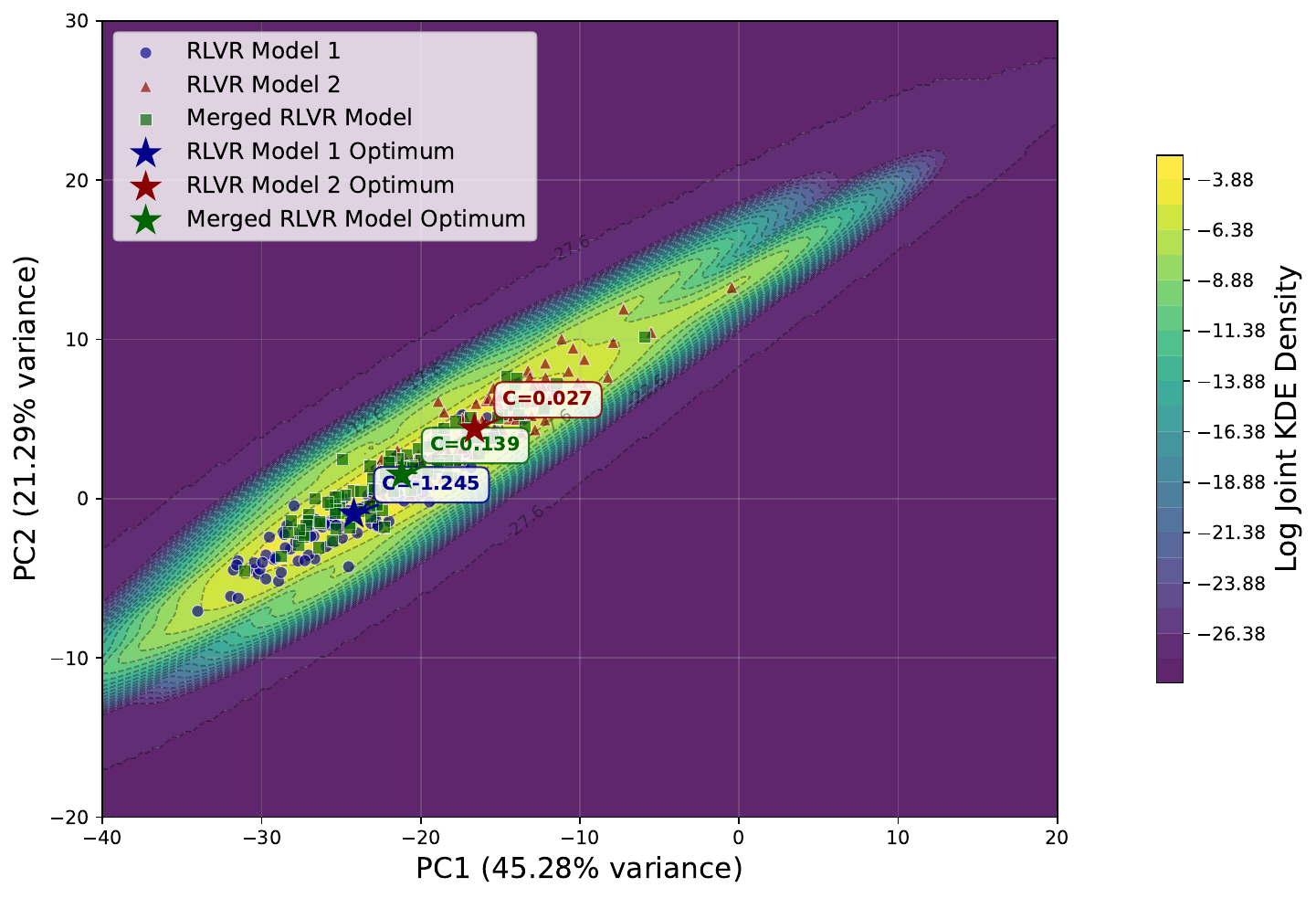}
     \label{fig:RL2D}}
     \subfigure[RL activation density (3D)]{
    \includegraphics[width=0.45\linewidth]{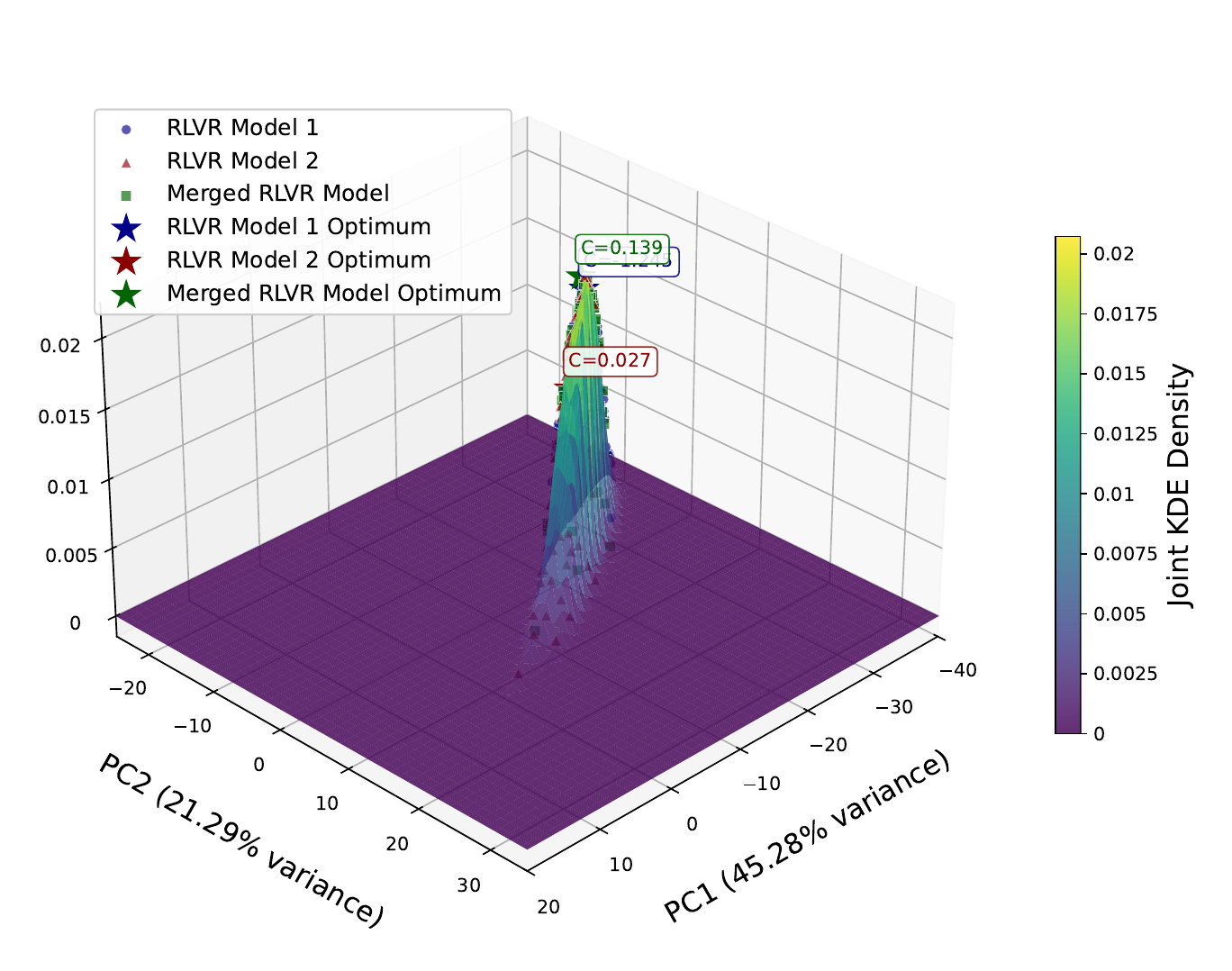}
     \label{fig:RL3D}}
     \subfigure[SFT activation density (2D)]{
\includegraphics[width=0.5\linewidth]{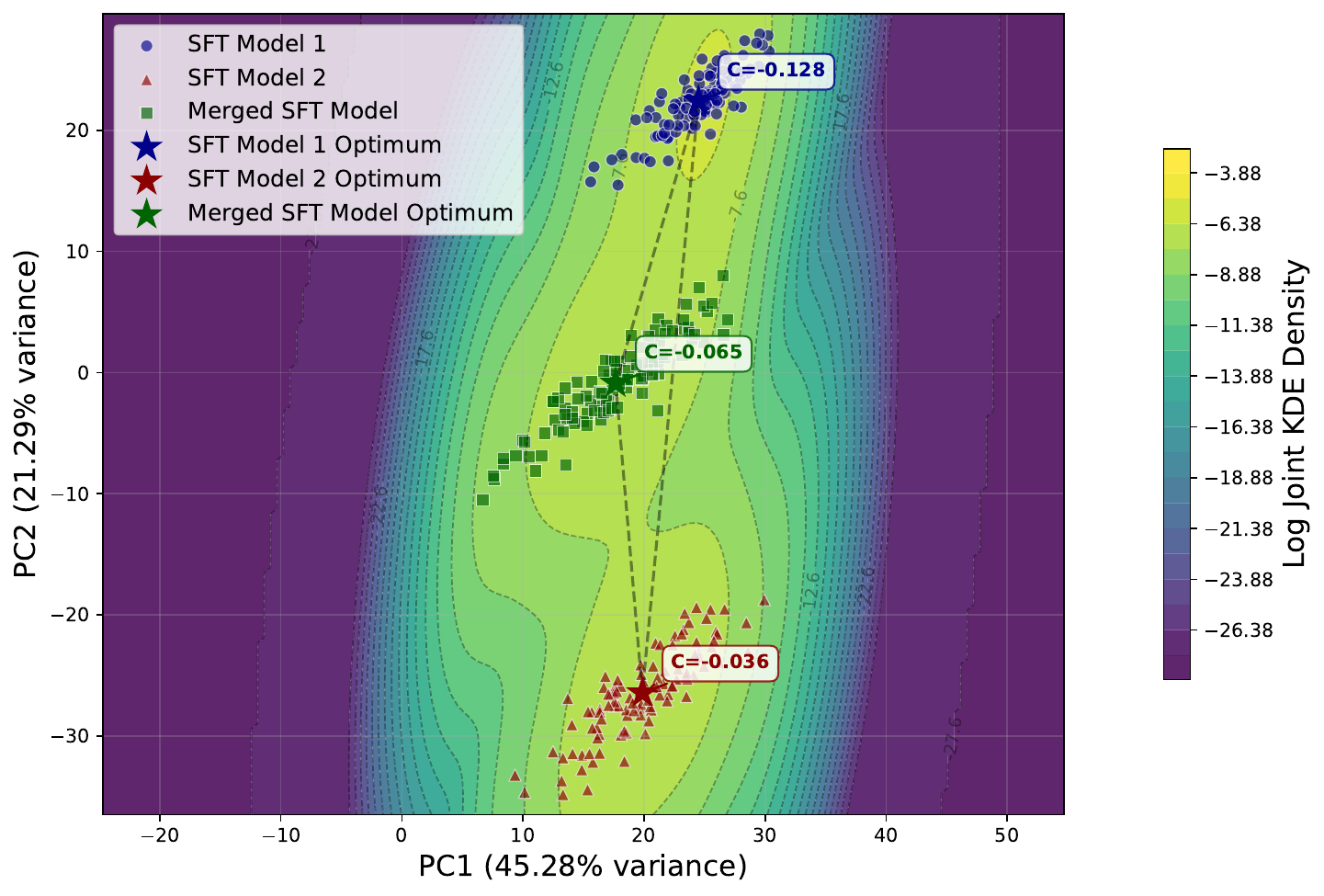}
     \label{fig:SFT2D}}
     \subfigure[SFT activation density (3D)]{
    \includegraphics[width=0.45\linewidth]{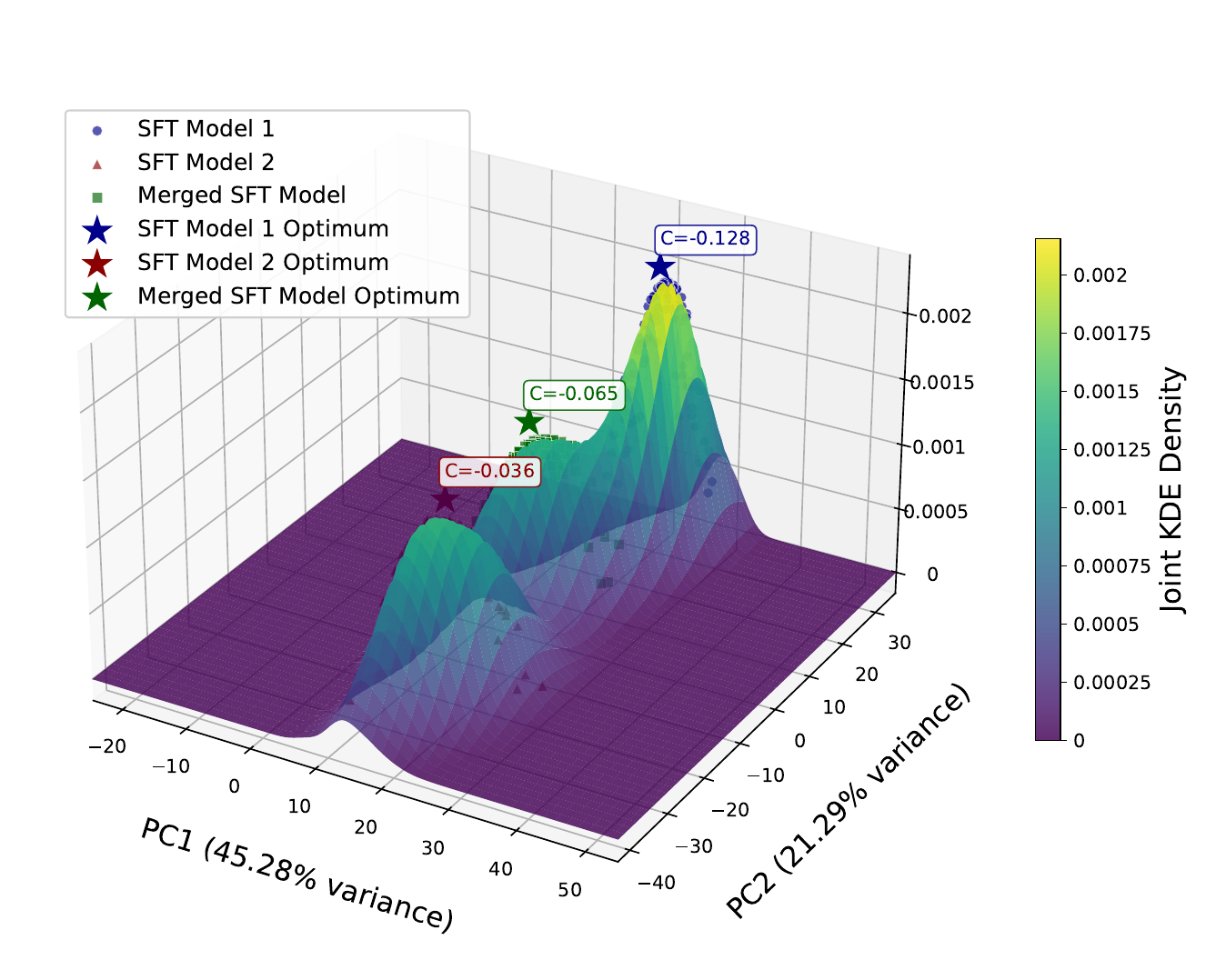}
     \label{fig:SFT3D}}
     \vspace{-0.2cm}
    \caption{Activation density landscape of 1.5B RL and SFT models. RL models exhibit sharp, isolated density peaks with abrupt transitions. SFT models share a smooth, connected density ridge across parent and merged models.}
    \label{fig:basin}
\end{figure*}
Based on the sparsity analysis of RLVR updates, we have observed that RLVR modifies the model in a highly selective manner, which differs substantially from SFT.
Previous work~\cite{zhu2025path} shows that SFT updates tend to move toward high-energy and high-curvature principal singular directions, while RL updates, constrained by KL anchoring and policy pruning, proceed through smaller and more localized steps.
However, directional analysis alone cannot fully explain why SFT models can often be merged effectively, whereas RLVR models suffer from severe degradation after merging.
To further understand this phenomenon, we analyze model merging from the perspective of activation-space geometry.

\textbf{Local geometry of activation density.}
We study the geometry of model \emph{activation distributions} on a fixed downstream task in Figure~\ref{fig:basin}.
For each training regime, including RLVR and SFT, we collect hidden activations from a fixed transformer layer over a shared set of GSM8K prompts for the two parent models and their averaged merged model.
The resulting activation samples $\{h_i^t\}_{i=1}^N \subset \mathbb{R}^d$ are projected into a shared two-dimensional latent space using PCA, yielding representations $\{z_i^t\}_{i=1}^N \subset \mathbb{R}^2$.
In the visualization, PC1 and PC2 explain 45.28\% and 21.29\% of the variance, respectively.
We then estimate a \emph{joint} smooth density $p_{\mathrm{joint}}(z)$ over the pooled projected activations from the parent and merged models via kernel density estimation (KDE).
Thus, the contours and 3D surfaces in Figure~\ref{fig:basin} characterize the shared activation support of a model family rather than the density of a single model.
High-density regions indicate activation configurations frequently occupied by at least one model, while low-density or unstable transition regions indicate poor shared support among the models.

To characterize the local geometry around each model on this joint landscape, we define a local patch around the model-specific activation center.
Let $z_t^\ast$ denote the representative activation point of model $t$, shown as the star marker in Figure~\ref{fig:basin}.
We construct a local patch:
\begin{equation}
    \mathcal{P}_t \;=\; \{ z \in \mathbb{R}^2 \mid \| z - z_t^\ast \|_\infty \le \epsilon \},
\end{equation}
where $\epsilon$ is a fixed radius shared across models.
This patch captures the local neighborhood around the activation region occupied by each model and supports curvature estimation through finite-difference approximations of the Hessian of the log joint density.

Formally, within each local patch $\mathcal{P}_t$, we quantify the local curvature by the trace of the Hessian of $\log p_{\mathrm{joint}}(z)$ evaluated at $z_t^\ast$:
\begin{equation}
    C_t \;\triangleq\;
    \mathrm{Tr}\!\left( \nabla^2_z \log p_{\mathrm{joint}}(z) \right)\Big|_{z = z_t^\ast},
\end{equation}
which is approximated numerically via finite differences over the patch.
Since $C_t$ is computed on the joint activation density rather than on a model-specific density, its sign is informative.
A negative $C_t$ indicates that the model lies in a locally concave high-density region; when $C_t<0$, a larger $|C_t|$ corresponds to a sharper and more isolated local peak.
In contrast, a near-zero or positive $C_t$ suggests a flat, saddle-like, or locally convex transition region on the joint landscape, implying that the corresponding activation point is not supported by a stable shared density basin.

For SFT models, Figure~\ref{fig:SFT2D} and Figure~\ref{fig:SFT3D} show a broad and connected activation landscape.
Although the two parent models occupy separated regions, mainly along PC2, their activation distributions are connected by a smooth high-density band.
The merged SFT model lies between the two parent models on this band rather than falling into a low-density gap.
Consistently, the local curvature estimates of the SFT models are all negative: $C=-0.128$ for SFT Model 1, $C=-0.036$ for SFT Model 2, and $C=-0.065$ for the merged SFT model.
The merged model is not the flattest point among the three, but it remains inside a locally concave and well-supported region of the joint density.
The 3D density surface further shows that the parent and merged activation regions are arranged along a continuous ridge.
This indicates that SFT models, even when their parameter updates differ, tend to map inputs onto a broad and approximately connected activation manifold.
As a result, parameter averaging can still place the merged SFT model in an in-distribution activation region, explaining why direct merging is often effective for SFT models.

In contrast, RLVR models exhibit a much narrower and more fragile activation geometry.
As shown in Figure~\ref{fig:RL2D}, the RLVR activations concentrate along a thin diagonal region, and the contours become tightly packed around the high-density path.
The 3D view in Figure~\ref{fig:RL3D} further reveals a sharp density surface with abrupt drops away from this narrow support.
The local curvature values are also highly asymmetric: RLVR Model 1 lies on a sharp locally concave peak with $C=-1.245$, while RLVR Model 2 has $C=0.027$ and the merged RLVR model has $C=0.139$.
The positive curvature of the merged model indicates that it does not lie in a stable concave basin of the joint activation landscape, but instead falls into a saddle-like or locally convex transition region.
Therefore, although the RLVR parent and merged activations may appear close in the two-dimensional projection, their shared activation support is much sharper and less tolerant to interpolation than that of SFT models.
This suggests that independent RLVR models develop fragile and partially incompatible activation pathways.
Consequently, naive parameter averaging can move the merged model away from a stable activation basin and into an unstable transition region, causing interference between RLVR-induced reasoning shortcuts.

\section{Methodology}
\subsection{Overview}
\begin{figure}[t]
    \centering
  \includegraphics[width=0.7\columnwidth]{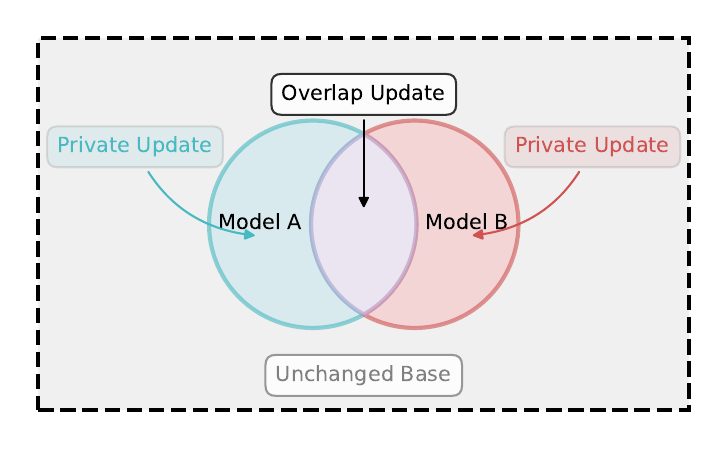}
      \vspace{-0.2cm}
    \caption{Illustration of sparse RLVR model merging in an area view. Divided into overlap, private, and unchanged areas. The overlap area has subspaces of symbolic conflict.}
    \label{fig:venn}
    \vspace{-0.5cm}
\end{figure}
The observations above highlight two key challenges for merging RLVR models. First, the sparsity of RL updates creates distinct overlapping regions during aggregation (Figure~\ref{fig:venn}), each requiring different handling. Second, independent RLVR models learn near-orthogonal update directions, unlike SFT models that converge toward shared regions. To address these challenges, we propose Sensitivity-aware Resolving Merging (SAR-Merging). SAR-Merging selectively resolves conflicts in overlapping update regions by prioritizing parameters that are most influential to model behavior. It further applies sparsity-preserving rescaling on the merged task vectors to prevent the homogenization that destroys RLVR reasoning pathways.

\subsection{Sensitivity Quantification}
We first analyze the overlap status of the two merging models. Let $\delta^A = \theta^A - \theta^0$ and $\delta^B = \theta^B - \theta^0$ denote the task vectors of two fine-tuned models. According to the update status of each parameter, we divide the parameter space into conflict, overlap, private, and unchanged regions, as shown in Figure~\ref{fig:venn}. The conflict region contains parameters updated by both models but in opposite directions:
\begin{equation}
    \mathcal{C} = \{j \mid \delta^A_j \delta^B_j < 0\}.
\end{equation}
The non-conflicting overlap region satisfies $\delta^A_j \delta^B_j > 0$, the private region contains parameters updated by only one model, and the unchanged region contains parameters updated by neither model.

We then quantify the importance in the conflict region $\mathcal{C}$. Let $\mathcal{D}_{\text{val}}$ be a small validation set (e.g., parts of samples from GSM8K). We compute the diagonal Fisher Information Matrix~\cite{fisher}, denoted as sensitivity vector $\mathcal{S}^t \in \mathbb{R}^d$, where the $j$-th element is estimated via the squared gradients of activation:
\begin{equation}
    \mathcal{S}^t_j 
    =
    \mathbb{E}_{(x,y)\sim \mathcal{D}_{\text{val}}}
    \left[
    \left(
    \frac{\partial \log p_{\theta}(y|x;\theta^t)}
    {\partial \theta^t_j}
    \right)^2
    \right],
    \quad j \in \mathcal{C}.
\end{equation}
Intuitively, a larger $\mathcal{S}^t_j$ implies that perturbing the $j$-th parameter has a stronger impact on the output distribution of model $t$, indicating higher task sensitivity. 
Therefore, in the conflict region, Fisher sensitivity provides a local criterion for deciding which model's update should be preserved.

\subsection{Conflict Resolution}

Given two task vectors $\delta^A$ and $\delta^B$ derived from models $\theta^A$ and $\theta^B$, we identify parameters where the models disagree. 
In the overlap area, a sign conflict occurs at index $j$ if the two updates move in opposite directions, i.e., $\delta^A_j \delta^B_j < 0$.
Existing methods~\cite{ties,DARE} often resolve conflicts by magnitude pruning, averaging, or sign voting. 
However, such voting policies empirically do not perform well for RLVR models, whose sparse updates may correspond to task-specific reasoning shortcuts. 
Therefore, we employ sensitivity-based arbitration only in the conflict region. 
The resolved task vector $\delta^r$ is defined element-wise as
\begin{equation}
    \delta^{r}_j = 
    \begin{cases} 
        \delta^A_j + \delta^B_j, 
        & \text{if } \delta^A_j \cdot \delta^B_j \ge 0, \\

        \delta^A_j, 
        & \text{if } \delta^A_j \cdot \delta^B_j < 0 
        \text{ and } \mathcal{S}^A_j > \mathcal{S}^B_j, \\

        \delta^B_j, 
        & \text{if } \delta^A_j \cdot \delta^B_j < 0 
        \text{ and } \mathcal{S}^B_j \ge \mathcal{S}^A_j.
    \end{cases}
\end{equation}
This mechanism ensures that in regions of interference, the update from the model with higher task sensitivity is preserved, effectively silencing the less confident update.
\subsection{Sparsification and Rescaling}

To further reduce noise and preserve the sparse structure of RLVR updates, we apply DARE-style sparsification~\cite{DARE} to low-magnitude parameters in the private update region. 
Let $M \in \{0,1\}^d$ be a random binary mask, where each sparsified entry is sampled as $M_j \sim \mathrm{Bernoulli}(1-p),$ and $p$ denotes the dropout rate. In practice, we apply dropout only to the lower-magnitude half of the private update region $\delta^r_{p}$, since these parameters are more likely to contain weak or noisy task-specific changes. For RLVR models, we generally adopt a smaller dropout rate than for SFT models to avoid damaging sparse reasoning paths. The final task vector is obtained by rescaling the retained parameters:
\begin{equation}
    \delta^{\text{final}}
    =
    \frac{1}{1-p} (M \odot \delta^r_{p}),
\end{equation}
where $\odot$ denotes the Hadamard product. 
The rescaling factor $(1-p)^{-1}$ compensates for the expected shrinkage caused by random sparsification. 
Specifically, for an entry subject to sparsification, the unscaled sparse update is $\delta'_j = M_j \delta^r_j,$ with expectation
$\mathbb{E}[\delta'_j]=\mathbb{E}[M_j]\delta^r_j=(1-p)\delta^r_j.$
After rescaling, we have $\mathbb{E}\left[\frac{1}{1-p}M_j\delta^r_j\right]=\delta^r_j.$ Therefore, the rescaling preserves the expected update magnitude after sparsification, providing a formal justification for using $(1-p)^{-1}$ across different sparsity levels.

Finally, the merged model parameters are obtained by injecting the processed task vector into the base model with a scaling coefficient $\lambda$:
\begin{equation}
    \theta_{\text{merge}}
    =
    \theta^0 + \lambda \delta^{\text{final}}.
\end{equation}

\begin{figure}[!t]
    \centering
         \subfigure[1.5B models]{
    \includegraphics[width=0.9\columnwidth]{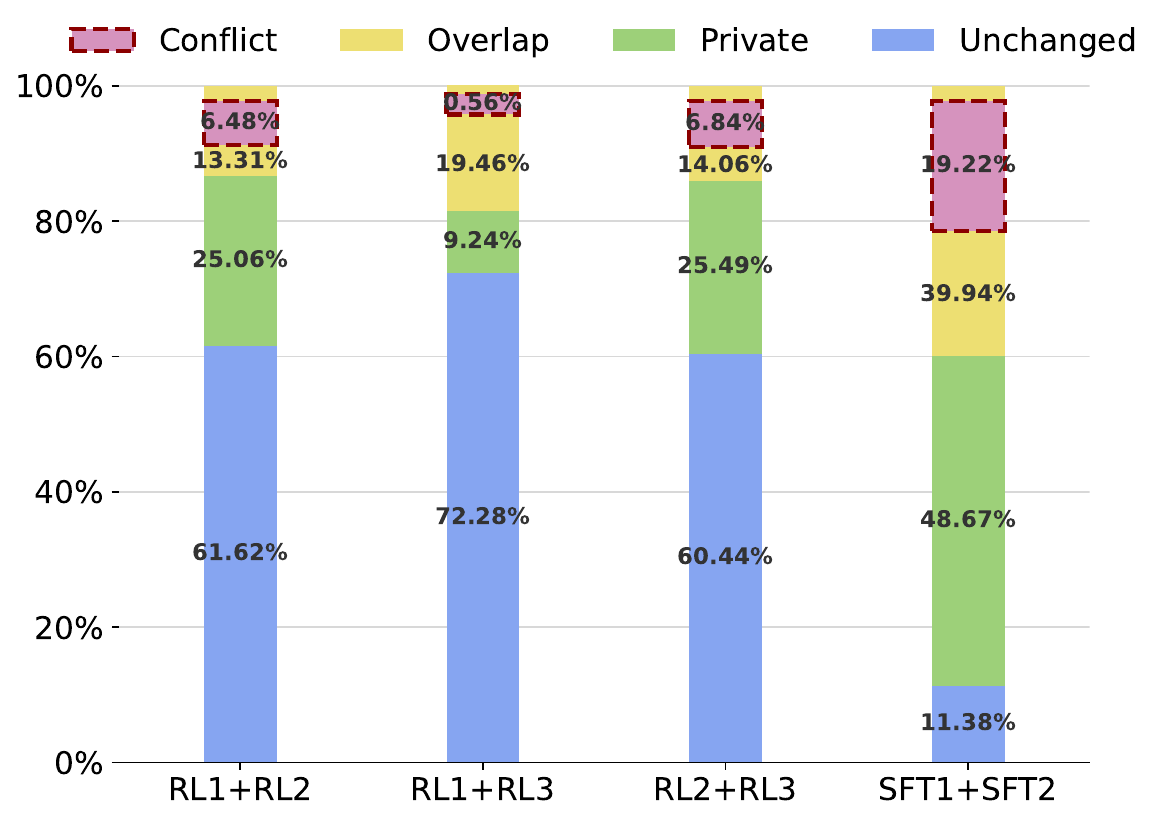}
     \label{fig:1.5Boverlap}}
         \subfigure[7B models]{
    \includegraphics[width=0.9\columnwidth]{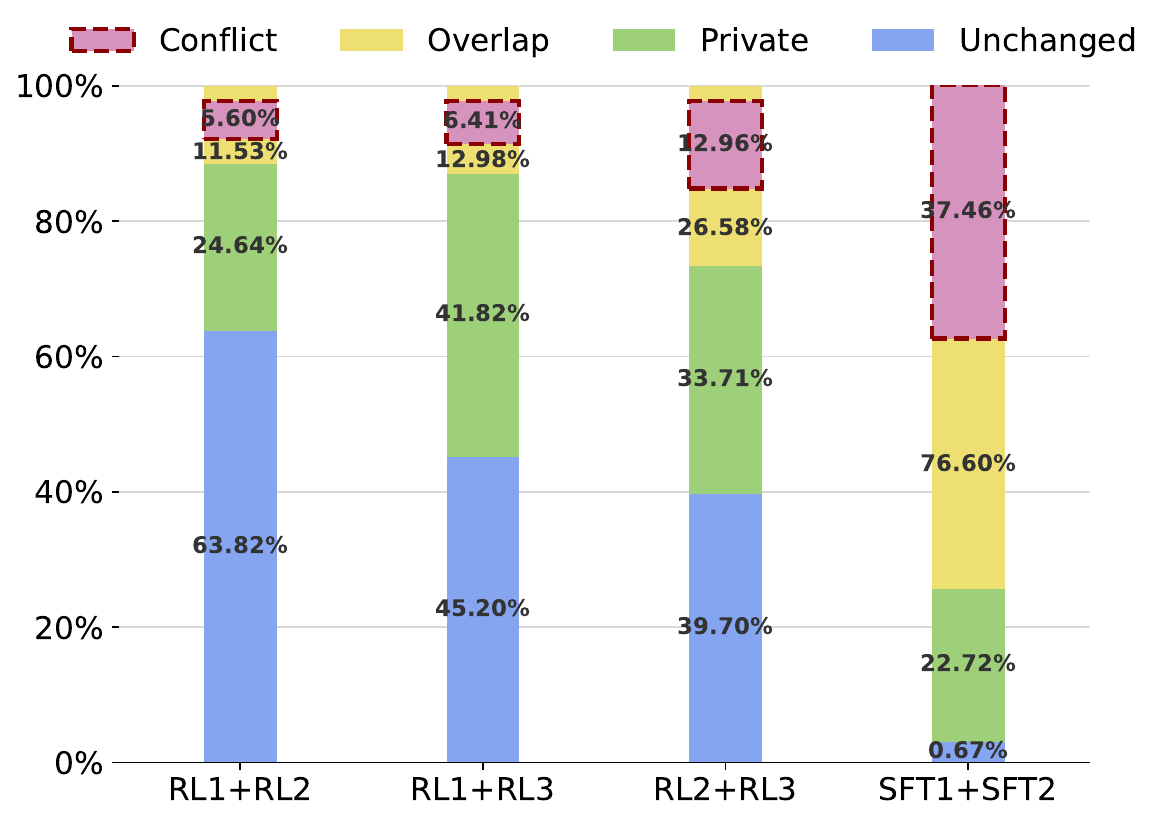}
     \label{fig:7Boverlap}}
    \caption{ Visualization of different update portions of two merged models of 1.5B and 7B. The conflict region is a subset of overlapping regions.}
    \label{fig:distribution}
   \vspace{-0.5cm}
\end{figure}

\setlength{\tabcolsep}{10pt}
\begin{table*}[!t]
\caption{Evaluations on 1.5B Math RLVR model merging. Parent model performances are shown below each benchmark name. (Gains $\uparrow$) are computed over Naive Avg. The \textbf{bold} and \underline{underline} fonts highlight the best and second-best methods. The accuracy of \textcolor{base}{Base} model DeepSeek-R1-Distill-Qwen-1.5B for GSM8K and MATH is 57.24\% and 33.92\%.}
\centering
\resizebox{\textwidth}{!}{
\begin{tabular}{@{}l|cc|cc|cc|cc@{}}
\toprule
\multirow{2}{*}{\textbf{Merging Method}}
& \multicolumn{2}{c|}{\shortstack{\textcolor{rl1}{RL$_1$} \& \textcolor{rl2}{RL$_2$}\\
{\scriptsize DeepScaleR-1.5B-Preview \& RLinf-math-1.5B}}}
& \multicolumn{2}{c|}{\shortstack{\textcolor{rl1}{RL$_1$} \& \textcolor{rl3}{RL$_3$}\\
{\scriptsize DeepScaleR-1.5B-Preview \& E1-Math-1.5B}}}
& \multicolumn{2}{c|}{\shortstack{\textcolor{rl2}{RL$_2$} \& \textcolor{rl3}{RL$_3$}\\
{\scriptsize RLinf-math-1.5B \& E1-Math-1.5B}}}
& \multicolumn{2}{c}{\shortstack{\textcolor{sft1}{SFT$_1$} \& \textcolor{sft2}{SFT$_2$}\\
{\scriptsize lul-sft \& MiniMath}}} \\
\cmidrule(lr){2-3}\cmidrule(lr){4-5}\cmidrule(lr){6-7}\cmidrule(l){8-9}
& \shortstack{\textbf{GSM8K}\\{(58.60, 67.17)}}
& \shortstack{\textbf{MATH}\\{(52.84, 67.50)}}
& \shortstack{\textbf{GSM8K}\\{(58.60, 67.47)}}
& \shortstack{\textbf{MATH}\\{(52.84, 62.96)}}
& \shortstack{\textbf{GSM8K}\\{(67.17, 67.47)}}
& \shortstack{\textbf{MATH}\\{(67.50, 62.96)}}
& \shortstack{\textbf{GSM8K}\\{(60.12, 76.57)}}
& \shortstack{\textbf{MATH}\\{(40.82, 57.20)}} \\
\midrule
Naive Avg.
& 62.88 & 60.17
& 63.03 & 57.90
& 67.32 & 65.23
& 68.34 & 49.01 \\
\midrule
Linear-Merging
& 55.64 {(-7.24)} & 47.34 {(-12.83)}
& 55.64 {(-7.39)} & 47.34 {(-10.56)}
& 55.42 {(-11.90)} & 48.34 {(-16.89)}
& 70.12 {(+1.78)} & \textbf{45.34} {(-3.67)} \\

Task Arithmetic
& \underline{65.42} {(+2.54)} & \underline{50.52} {(-9.65)}
& 56.78 {(-6.25)} & 50.08 {(-7.82)}
& \textbf{71.87} {(+4.55)} & \underline{51.80} {(-13.43)}
& 60.27 {(-8.07)} & 34.60 {(-14.41)} \\

TIES-Merging
& 65.39 {(+2.51)} & 48.10 {(-12.07)}
& 56.33 {(-6.70)} & \underline{52.92} {(-4.98)}
& 61.94 {(-5.38)} & 49.64 {(-15.59)}
& \textbf{70.50} {(+2.16)} & \underline{42.08} {(-6.93)} \\

DARE-Merging
& 60.57 {(-2.31)} & 48.28 {(-11.89)}
& \underline{57.71} {(-5.32)} & 51.42 {(-6.48)}
& 58.75 {(-8.57)} & 50.26 {(-14.97)}
& 69.74 {(+1.40)} & 41.98 {(-7.03)} \\

RAM-Merging
& 60.80 {(-2.08)} & 49.64 {(-10.53)}
& 54.43 {(-8.60)} & 51.56 {(-6.34)}
& 65.20 {(-2.12)} & 50.70 {(-14.53)}
& / & /\\

\textbf{Ours}
& \textbf{68.34} {(+5.46)} & \textbf{58.70} {(-1.47)}
& \textbf{66.75} {(+3.72)} & \textbf{59.32} {(+1.42)}
& \underline{70.46} {(+3.14)} & \textbf{59.52} {(-5.71)}
& / & / \\
\bottomrule
\end{tabular}}
\label{table:1.5Bmain}
\end{table*}

\section{Experiments}
\subsection{Experimental Settings}
\textbf{Models and Datasets.} We conduct evaluations on four reasoning datasets: GSM8K~\cite{gsm8k} and MATH~\cite{math} for mathematical testing. We adopt HumanEval~\cite{human_eval} and MBPP~\cite{mbpp} for code generation evaluation. We use the models shown in Table~\ref{tab:sparsity_joint}. We compare 1.5B math RLVR models based on the open-source DeepSeek-R1-Distill-Qwen-1.5B~\cite{GRPO} for the mathematical task, where we also add SFT models as reference. We evaluate 1.5B math and coder model merging. We also test 7B math RLVR models with the same basis on open-source Qwen2.5-Math-7B~\cite{Qwen2.5-Math}, compared with SFT models.

\textbf{Evaluation Metrics.} We calculate zero-shot accuracy for GSM8K and MATH with classical answer extraction methods. For the code task, we adopt pass@1 for HumanEval and MBPP.

\textbf{Baselines.} To evaluate our method, we compare with five classical model merging methods: Linear-Merging~\cite{modelsoup}, Task Arithmetic~\cite{TaskArithmetic}, TIES-Merging~\cite{ties}, DARE-Merging~\cite{DARE}, and RAM-Merging~\cite{ram} for RL model merging.

\textbf{Implementation Details.} We apply dropout only for methods that require it. For DARE-Merging, we use dropout rates of 0.2 and 0.9 for the 1.5B and 7B models, respectively, with averaged merging for fairness. For RAM-Merging, we adopt the original hyperparameters with
$r = 0.1$ and $\alpha= 2.0$. We set the dropout rate as 0.2 for our method. As stated in Table~\ref{tab:sparsity_joint}, all models are operated in bfloat16 with threshold value of $1\times10^{-5}$. Experiments are conducted on NVIDIA RTX 5090 GPUs.

\subsection{Main Results}
\setlength{\tabcolsep}{10pt}
\begin{table*}[t]
\caption{Evaluations on 7B Math RLVR model merging. Parent model performances are shown below each benchmark name. (Gains $\uparrow$) are computed over Naive Avg. The \textbf{bold} and \underline{underline} fonts highlight the best and second-best methods. The accuracy of \textcolor{base}{Base} model Qwen-2.5-7B-MATH for GSM8K and MATH is 85.97\% and 71.48\%.}
\centering
\vspace{-0.3cm}
\resizebox{\textwidth}{!}{
\begin{tabular}{@{}l|cc|cc|cc|cc@{}}
\toprule
\multirow{2}{*}{\textbf{Merging Method}}
& \multicolumn{2}{c|}{\shortstack{\textcolor{rl1}{RL$_1$} \& \textcolor{rl2}{RL$_2$}\\
{\scriptsize Qwen-2.5-7B-MATH-RL \& Qwen2.5-Math-7B-Oat-Zero}}}
& \multicolumn{2}{c|}{\shortstack{\textcolor{rl1}{RL$_1$} \& \textcolor{rl3}{RL$_3$}\\
{\scriptsize Qwen-2.5-7B-MATH-RL \& Qwen2.5-Math-7B-GPG}}}
& \multicolumn{2}{c|}{\shortstack{\textcolor{rl2}{RL$_2$} \& \textcolor{rl3}{RL$_3$}\\
{\scriptsize Qwen2.5-Math-7B-Oat-Zero \& Qwen2.5-Math-7B-GPG}}}
& \multicolumn{2}{c}{\shortstack{\textcolor{sft1}{SFT$_1$} \& \textcolor{sft2}{SFT$_2$}\\
{\scriptsize Aryabhata-1.0 \& Satori-SFT-7B}}} \\
\cmidrule(lr){2-3}\cmidrule(lr){4-5}\cmidrule(lr){6-7}\cmidrule(l){8-9}
& \shortstack{\textbf{GSM8K}\\{(87.33, 90.82)}}
& \shortstack{\textbf{MATH}\\{(72.54, 80.06)}}
& \shortstack{\textbf{GSM8K}\\{(87.33, 84.23)}}
& \shortstack{\textbf{MATH}\\{(72.54, 77.56)}}
& \shortstack{\textbf{GSM8K}\\{(90.82, 84.23)}}
& \shortstack{\textbf{MATH}\\{(80.06, 77.56)}}
& \shortstack{\textbf{GSM8K}\\{(86.80, 86.17)}}
& \shortstack{\textbf{MATH}\\{(71.52, 45.54)}} \\
\midrule
Naive Avg.
& 89.07 & 76.30
& 85.78 & 75.05
& 87.52 & 78.81
& 86.48 & 58.53 \\
\midrule
Linear-Merging
& 82.41 {(-6.66)} & 63.02 {(-13.28)}
& 83.01 {(-2.77)} & 63.34 {(-11.71)}
& 78.39 {(-9.13)} & \underline{60.84} {(-17.97)}
& \underline{94.16} {(+7.68)} & \underline{71.92} {(+13.39)} \\

Task Arithmetic
& 78.69 {(-10.38)} & 54.28 {(-22.02)}
& 80.59 {(-5.19)} & \underline{66.00} {(-9.05)}
& 78.08 {(-9.44)} & 46.66 {(-32.15)}
& 92.03 {(+5.55)} & 71.26 {(+12.73)} \\

TIES-Merging
& 74.90 {(-14.17)} & 55.04 {(-21.26)}
& 81.88 {(-3.90)} & 62.88 {(-12.17)}
& 74.82 {(-12.70)} & 57.34 {(-21.47)}
& 90.90 {(+4.42)} & 71.14 {(+12.61)} \\

DARE-Merging
& \underline{83.16} {(-5.91)} & \underline{63.42} {(-12.88)}
& \underline{83.47} {(-2.31)} & 62.70 {(-12.35)}
& \underline{80.36} {(-7.16)} & 60.82 {(-17.99)}
& \textbf{95.45} {(+8.97)} & \textbf{72.34} {(+13.81)} \\

RAM-Merging
& 73.01 {(-16.06)} & 53.48 {(-22.82)}
& 81.80 {(-3.98)} & 62.78 {(-12.27)}
& 74.14 {(-13.38)} & 60.34 {(-18.47)}
& / & / \\

\textbf{Ours}
& \textbf{86.59} {(-2.48)} & \textbf{70.78} {(-5.52)}
& \textbf{87.31} {(+1.53)} & \textbf{68.50} {(-6.55)}
& \textbf{83.62} {(-3.90)} & \textbf{68.56} {(-10.31)}
& / & / \\
\bottomrule
\end{tabular}}
\label{table:7Bmain}
\vspace{-0.3cm}
\end{table*}
\textbf{Results on 1.5B models of math tasks.}
We investigate the merging performance of 1.5B math models in Table~\ref{table:1.5Bmain}. When merging RLVR models, most existing merging baselines still suffer from clear performance degradation, especially on the more challenging MATH benchmark. For example, although Task Arithmetic achieves competitive performance on GSM8K in several cases, its MATH accuracy remains substantially below the naive average. In contrast, our SAR-Merging consistently achieves the best overall performance across all three RL model pairs. For RL$_1$ \& RL$_2$, SAR-Merging improves GSM8K from the naive average of 62.88\% to 68.34\%, and also substantially reduces the MATH degradation, achieving 58.70\% compared with 47.34\%--50.52\% of other merging baselines. We also examine the parameter overlap patterns across model pairs, as shown in Figure~\ref{fig:1.5Boverlap}. We found that higher conflict generally leads to a more difficult merging case: RL$_1$ \& RL$_3$ has the lowest conflict rate of 0.56\% and is the only pair where SAR-Merging improves over the naive average on both datasets, while RL$_1$ \& RL$_2$ and RL$_2$ \& RL$_3$ have higher conflict rates of 6.48\% and 6.84\%, respectively. This indicates that SAR-Merging is robust under moderate conflict, but the remaining MATH degradation is still related to conflicting RLVR updates.

\textbf{Results on 7B models of math tasks.}
As shown in Table~\ref{table:7Bmain}, the fusion of 7B models presents a more challenging setting for RLVR model merging. 
For RL model pairs, most existing merging methods significantly underperform the naive average on both GSM8K and MATH. By comparison, SAR-Merging achieves 86.59\% on GSM8K and 70.78\% on MATH, reducing the degradation to only -2.48\% and -5.52\%, respectively. A similar trend is observed for RL$_1$ \& RL$_3$, where SAR-Merging is the only method that surpasses the naive average on GSM8K, achieving 87.31\% with a gain of 1.53\%. The overlap distributions in Figure~\ref{fig:7Boverlap} show a similar pattern as 1.5B models: RL$_1$ \& RL$_2$ has the lowest conflict and overlap rates, 5.60\% and 11.53\%, and correspondingly exhibits the smallest degradation after SAR-Merging, while RL$_2$ \& RL$_3$ has the highest conflict and overlap rates, leading to the most difficult merging scenario. These results suggest that SAR-Merging can mitigate the negative impact of conflicting RLVR updates, but higher conflict and denser overlap still make preserving both models' capabilities more challenging.

\textbf{Results on 1.5B math \& coder models.} Multi-capability fusion is a central goal of model merging. As shown in Table~\ref{table:mathcode}, even with an extremely low parameter conflict rate of only 1.92\%, traditional fusion methods still suffer significant performance losses, particularly on coding tasks. For example, Linear-Merging drops HumanEval accuracy from 36.58\% (parent coder model) to 15.24\%, and DARE and RAM drop to 9.75\% and 14.63\%, respectively. These methods nearly destroy the model's programming capabilities. In contrast, our method demonstrates a significant advantage in preserving multi-task capabilities. On the GSM8K math task, our method achieves an accuracy of 58.06\%. We hypothesize that code generation is typically extremely sensitive to model parameters. Traditional averaging or pruning methods are prone to disrupting the sparse activation paths required by the code logic, especially the parameters from RLVR.

\setlength{\tabcolsep}{4pt}
\begin{table}[t]
\caption{Evaluations on 1.5B Math and Coder RLVR model merging of various methods. The \underline{underline}/\textbf{bold} fonts highlight the best baseline/our method. The models' Conflict/ Overlap/ Private/ Unchanged is 1.92\%/ 18.45\%/ 15.41\%/ 66.12\%.  }
\centering
\resizebox{1\linewidth }{!}{
\begin{tabular}{@{}c|c|c|c|c|c}
\toprule
\textbf{Model} &\textbf{Merging
Methods} &\textbf{GSM8K}&\textbf{MATH} &\textbf{HumanEval} &\textbf{mbpp} \\
\midrule
Math: DeepScaleR-1.5B-Preview & / &58.60\%& 52.84\%&32.92\%&28.80\%
\\
Coder: DeepCoder-1.5B-Preview  & / &4.76\%& 57.32\% &36.58\%&30.00\%\\
\midrule
\multirow{6}{*}{Math \& Coder}
& Linear-Merging & \underline{55.95}\% & 50.10\%& 15.24\%&10.60\%   \\
& Task Arithmetic &48.14\%&49.04\% &11.58\% &9.20\%  \\
& TIES-Merging &51.93\% &51.10\% &\underline{16.46}\%&10.80\% \\
& DARE-Merging & 55.11\%&50.48\%&9.75\%&11.60\%  \\
& RAM-Merging & 52.23\% &\underline{51.20}\% &14.63\% &\underline{12.00}\% \\
& \textbf{Ours} &\textbf{58.06}\% &\textbf{55.81}\% &\textbf{30.94}\% &\textbf{22.40}\% \\
\bottomrule
\end{tabular}
}
\label{table:mathcode}
\end{table}

\subsection{Ablation study}
We conduct an ablation study of SAR-Merging modules in Table~\ref{table:ablation}. All three components contribute positively, with sensitivity-based conflict resolution playing the most important role. From a geometric perspective, highly sensitive parameters correspond to directions with greater curvature in the loss landscape. Without sensitivity-based preservation, the fusion process can easily disrupt these critical parameters, causing the merged model to fall into high-loss regions. Sparsification and rescaling complement each other by maintaining the sparse structure of the merged task vector.

\setlength{\tabcolsep}{15pt}
\begin{table}[h]
\centering
\caption{Ablation study on math datasets of SAR-merging in terms of accuracy. Under RL$_1$ \& RL$_2$ of 7B math models.}
\resizebox{1\linewidth}{!}{%
 \begin{tabular}{ccc|cc}
\toprule 
Sensitivity & Sparsification   & Rescaling  & GSM8K & MATH \\
\midrule
\Checkmark & \Checkmark & - 
& 85.31\% & 68.42\% \\
\Checkmark & - & \Checkmark
& 84.76\% & 67.84\% \\
- & \Checkmark & \Checkmark
& 82.94\% & 63.52\% \\
\midrule
\Checkmark & \Checkmark & \Checkmark 
& \textbf{86.59}\% & \textbf{70.78}\% \\
\bottomrule
\end{tabular}
}
\label{table:ablation}
    \vspace{-0.4cm}
\end{table}

\section{Limitations and Future Work}
While SAR-Merging substantially reduces the degradation in RLVR model merging, a fundamental performance barrier persists in most settings: the merged model still underperforms the better individual parent model. In contrast, SFT model merging routinely produces merged models that surpass both parents. Closing this gap and enabling true reasoning aggregation, where merged RLVR models consistently outperform any single parent, remains an important open challenge. We believe that overcoming this barrier could unlock the full potential of combining diverse reasoning capabilities through merging.

Our experiments focus on the open-source Qwen and DeepSeek model families, and applicability to other architectures such as Llama requires further investigation. Finding paired SFT and RLVR models trained on the same base also remains challenging, limiting controlled comparisons. Our observations and methods need further validation on larger models (14B and 70B), where the interplay between sparsity and scale may differ.

Additionally, this work primarily considers pairwise merging of two models. Multi-model merging, which simultaneously aggregates three or more RLVR models with potentially diverse reasoning strategies, is a promising but largely unexplored direction. The conflict resolution and sparsity dynamics become significantly more complex with more models, and developing scalable methods for this setting is an important avenue for future research.

\section{Conclusion}
In this paper, we investigate the performance degradation observed when applying traditional model merging techniques to Large Language Models trained with Reinforcement Learning with Verifiable Reward (RLVR). Our empirical analysis reveals that, unlike Supervised Fine-Tuning, RL updates are inherently sparse and tend to form orthogonal shortcuts in parameter space, leading to interference during dense aggregation. To address this sparsity curse, we propose Sensitivity-aware Resolving Merging (SAR-Merging), a framework that prioritizes critical parameters using Fisher Information and enforces sparsity through rescaling. Experimental results on mathematical and coding benchmarks demonstrate that SAR-Merging significantly narrows the performance gaps compared to state-of-the-art baselines, effectively enabling multi-functional model fusion while preserving fragile reasoning capabilities.

\section*{Acknowledgments}
This work was supported by the National Key Research and Development Program of China (2025YFC2708700), Zhejiang Province Major Science and Technology Plan Project (2026C01021), the Central Government-Guided Fund for Local Science and Technology Development (2025ZY01034), the Provincial Key Laboratory Program (2025E10048), and Zhejiang Province Science and Technology Plan Project (2025C01128). Thanks to Wenhao Zhang for his discussion in the early stage of this work.


\newpage
\bibliographystyle{ACM-Reference-Format}
\balance
\bibliography{sample-base}

\appendix


\end{document}